\documentclass{article} 
\usepackage{eacl2017}
\usepackage{times}
\usepackage{latexsym}
\usepackage[draft]{hyperref}
\usepackage{url}
\usepackage{amsmath,amsthm,amsfonts}
\usepackage{multirow}
\usepackage{xspace}
\usepackage{tikz}
\usetikzlibrary{shapes,backgrounds,patterns}
\usepackage{graphicx}
\usepackage{caption}
\usepackage{subcaption}

\usepackage{algorithmicx}
\usepackage{algorithm}
\usepackage{algpascal}
\usepackage{algc}
\usepackage{algcompatible}
\usepackage{algpseudocode}
\usepackage{tabularx}

\newcommand{\citep}[1]{\cite{#1}}
\newcommand{\citet}[1]{\newcite{#1}}

\title{Generalizing to Unseen Entities and Entity Pairs with \\ Row-less Universal Schema}

\author{Patrick Verga, Arvind Neelakantan, \& Andrew McCallum \\
   College of Information and Computer Sciences\\
   University of Massachusetts Amherst\\
   \texttt{\{pat, arvind, mccallum\}@cs.umass.edu} \\
}

\begin{document}

\maketitle

\begin{abstract}
Universal schema predicts the types of entities and relations in a knowledge base (KB) by jointly embedding the union of all available schema types---not only types from multiple structured databases (such as Freebase or Wikipedia infoboxes), but also types expressed as textual patterns from raw text.  
This prediction is typically modeled as a matrix completion problem, with one type per column, and either one or two entities per row (in the case of entity types or binary relation types, respectively).  
Factorizing this sparsely observed matrix yields a learned vector embedding for each row and each column.  
In this paper we explore the problem of making predictions for entities or entity-pairs unseen at training time (and hence without a pre-learned row embedding).  
We propose an approach having no per-row parameters at all; rather we produce a row vector on the fly using a learned aggregation function of the vectors of the observed columns for that row.  
We experiment with various aggregation functions, including neural network attention models.  
Our approach can be understood as a natural language database, in that questions about KB entities are answered by attending to textual or database evidence.  
In experiments predicting both relations and entity types, we demonstrate that despite having an order of magnitude fewer parameters than traditional universal schema, we can match the accuracy of the traditional model, and more importantly, we can now make predictions about unseen rows with nearly the same accuracy as rows available at training time.
\end{abstract}


\section{Introduction\label{introduction}}

Automatic knowledge base construction (AKBC) is the task of building a structured knowledge base (KB) of facts using raw text evidence, and often an initial seed KB to be augmented~\citep{NELL,yago,freebase}. KBs generally contain entity type facts such as \emph{Sundar Pichai} \emph{IsA} \emph{Person} and relation facts such as \emph{CEO\_Of(Sundar Pichai, Google)}. Extracted facts about entities, and their types and relations are useful for many downstream tasks such as question answering~\citep{bordes2014question} and semantic parsing~\citep{dcs,ccg}.

An effective approach to AKBC is universal schema, which predicts the types of entities and relations in a knowledge base (KB) by jointly embedding the union of all available schema types---not only types from multiple structured databases (such as Freebase or Wikipedia infoboxes), but also types expressed as textual patterns from raw text. This prediction is typically modeled as a matrix completion problem. In the standard formulation for relation extraction~\citep{limin}, entity pairs and relations occupy the rows and columns of the matrix respectively (Figure \ref {fig:uschema-matrix}a). Analogously in entity type prediction~\citep{yao2013universal}, entities and types occupy the rows and columns of the matrix respectively (Figure \ref{fig:uschema-matrix}b).  The row and column entries are represented as learned vectors  with compatibility determined by a scoring function.

In its original form, universal schema can reason only about row entries and column entries explicitly seen during training. Unseen rows and columns observed at test time do not have a learned embedding. This problem is referred to as the \emph{cold-start} problem in recommendation systems~\citep{schein2002methods}.

Recently \newcite{toutanova2015representing} and \newcite{verga2015multilingual} proposed `column-less' versions of universal schema models that generalize to unseen column entries.  They    learn compositional pattern encoders to parameterize the column matrix in place of individual column embeddings. However, these models still do not generalize to unseen row entries. 

In this work, we present a `row-less' extension of universal schema that generalizes to unseen entities and entity pairs. Rather than representing each row entry with an explicit dense vector, we encode each entity or entity pair as aggregate functions over their observed column entries. This is beneficial because when new entities are mentioned in text documents and subsequently added to the KB, we can directly reason on the observed text evidence to infer new binary relations and entity types for the new entities. This avoids the cumbersome effort of re-training the whole model from scratch to learn embeddings for the new entities.

To construct the row representation, we compare various aggregation functions in our experiments. We consider query independent and dependent aggregation functions. We find that query dependent attentional models that selectively focus on relevant evidence outperform the query independent alternatives. The query dependent attention mechanism also helps in providing a direct connection between the prediction and its provenance. Additionally, our models have a much smaller memory footprint since they do not store explicit row representations. 

It is important to note that our approach is different from sentence level classifiers that predict KB relations and entity types using a single sentence as evidence. First, we pool information from multiple pieces of evidence coming from both text and annotated KB facts, rather than considering a single sentence at test time. Second, our methods are not limited to a fixed schema but instead predict a richer set of labels (KB types and textual), enabling easier downstream processing closer to natural language interaction with the KB. Finally, our model gains additional training signal from multi-task learning of textual and KB types. Since universal schema leverages large amounts of unlabeled text we desire the benefits of entity pair modeling, and row-less universal schema facilitates learning entity pair representations without the drawbacks of the traditional one-embedding-per-pair approach. 

The majority of current embedding methods for KB entity type prediction operate with explicit entity representations~\citep{yao2013universal,neelakantan_type} and hence, cannot generalize to unseen entities.  In relation extraction, entity-level models ~\citep{rescal,DBLP:journals/corr/Garcia-DuranBUG15,bishan,transe,wang2014knowledge,lin2015learning,socherkb} can handle unseen entity pairs at test time. These models learn representations for the entities instead of entity pairs. Hence, these methods still cannot generalize to predict relations between an entity pair if even one of the entities is unseen. Moreover, ~\citet{toutanova2015representing} and~\citet{limin} observe that the entity pair model outperforms entity models in cases where the entity pair was seen at training time. 

Most similar to this work, \citet{neelakantan2015compositional} classify KB relations by finding the maximum scoring path between two entities.
This model is also `row-less' and does not directly model entities or entity pairs.
There are several important differences in this work.
\citet{neelakantan2015compositional} learn per-relation classifiers to predict only a small set of KB relations, while we instead predict all relations, including textual relations.
We also explore aggregation functions that pool evidence from multiple paths while \citet{neelakantan2015compositional} only chose the maximum scoring path.
Additionally, we demonstrate that our models can perform on par with those with explicit row representations while \citet{neelakantan2015compositional} did not perform this comparison.

In this paper we investigate universal schema models without explicit row representations on two tasks: entity type prediction and relation extraction.
We use entity type and relation facts from Freebase~\citep{freebase} augmented with textual relations and types from Clueweb text~\citep{clueweb,gabrilovich2013facc1}. We explore multiple aggregation functions and find that an attention-based aggregation function outperforms several simpler functions and matches a model using explicit row representations with an order of magnitude fewer parameters.
More importantly, we then demonstrate that our `row-less' models accurately predict relations on unseen entity pairs and types on unseen entities. 


\section {Background: Universal Schema}
\begin{figure}[h]
\begin{minipage}{.45\textwidth}
\centering
\includegraphics[scale=.33]{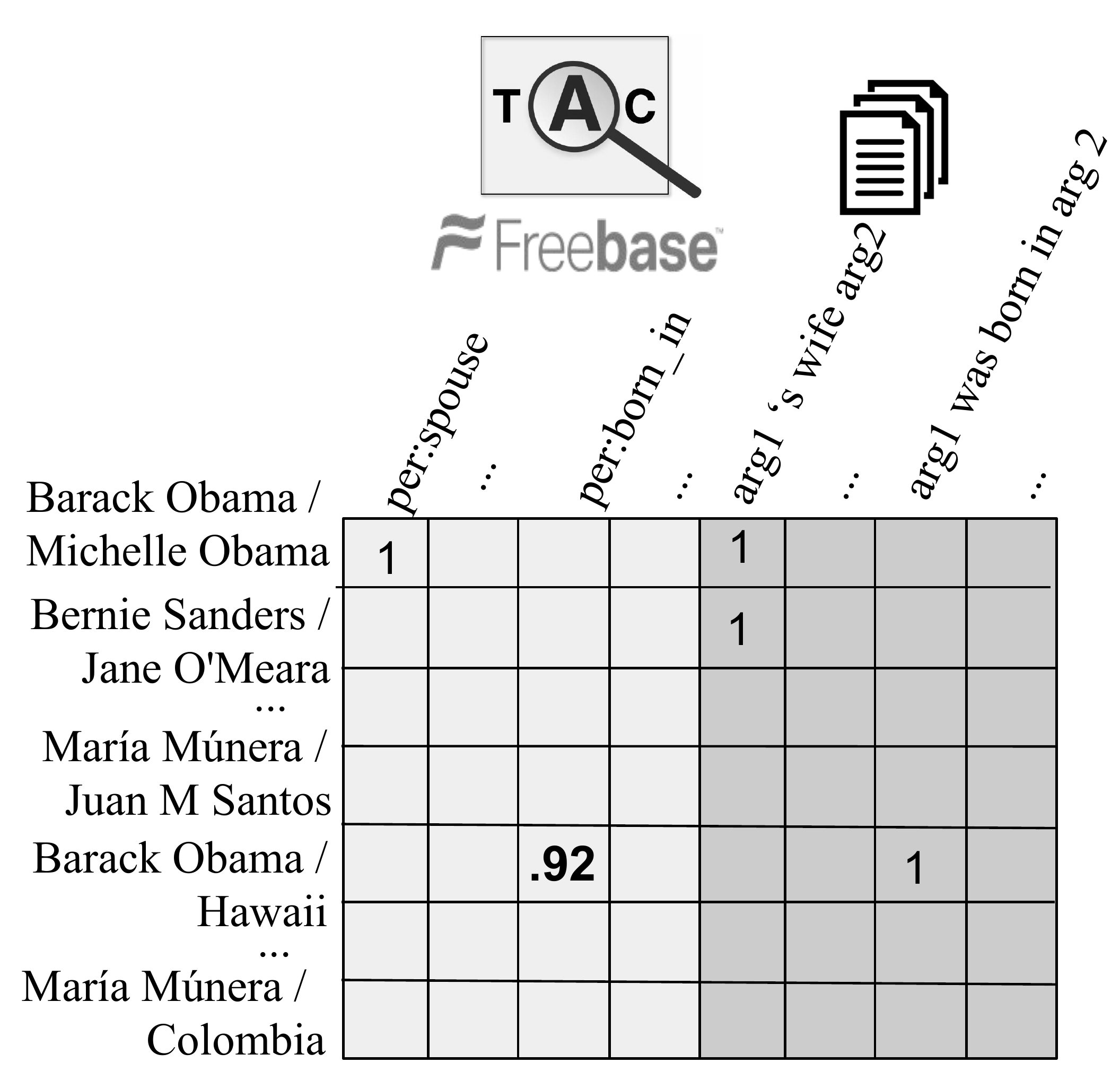}
\end{minipage}
\hspace{.5cm}
\begin{minipage}{.45\textwidth}
\centering
\includegraphics[scale=.65]{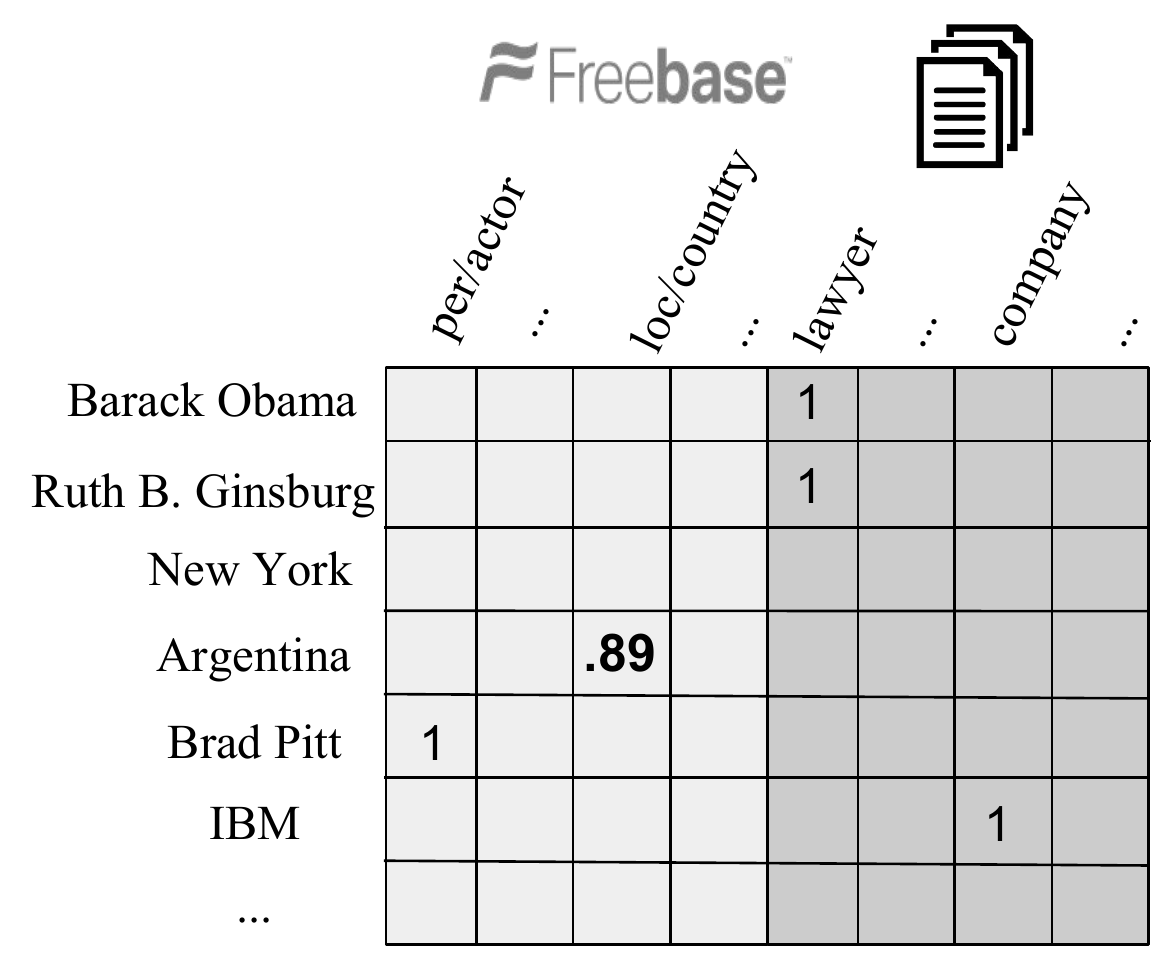}
\end{minipage}
\caption{Universal schema matrix. a: Relation extraction.
Relation types are represented as columns and entity pairs as rows of a matrix.
Both KB relation types and textual patterns from raw text are jointly embedded in the same space.
b: Entity type prediction.
Entity types are represented as columns and entities as rows of a matrix.
\label{fig:uschema-matrix}}
\end{figure}

Universal schema \citep{limin,yao2013universal} relation extraction and entity type prediction is typically  modeled as a matrix completion task. In relation extraction, entity pairs and relations occupy the rows and columns of the matrix (Figure \ref {fig:uschema-matrix}-a), while in entity type prediction, entities and types occupy the rows and columns of the matrix (Figure \ref{fig:uschema-matrix}-b). During training, we observe some positive entries in the matrix and at test time, we predict the missing cells in the matrix. This is  achieved by decomposing the observed matrix into two low-rank matrices resulting in embeddings for each column entry and each row entry. Test time prediction is performed using the learned low-rank column and row representations.

Let $T$ be the training set consisting of examples of the form $(r, c)$, where row $r \in U$ and column $c \in V$, denote an entity pair and relation type in the relation extraction task, or entity and entity type in the entity type prediction task. Let $v(r) \in \mathbb{R}^d$ and $v(c) \in \mathbb{R}^d$ be the vector representations or embeddings of row $r \in U$ and column $c \in V$ that are learned during training. Given a positive example, $(r,c) \in T$ in training, the probability of observing the fact is given by,
\begin{equation}
P(y_{r,c}=1) = \sigma(v(r).v(c))
\label{prob}
\end{equation}
where $y_{r,c}$ is a binary random variable that is equal to $1$ when $(r,c)$ is a fact  and $0$ otherwise, and $\sigma$ is the sigmoid function.
The embeddings are learned using Bayesian Personalized Ranking (BPR)~\citep{rendle2009bpr} in which the probability of the observed triples are ranked above unobserved triples.

\section{Model \label{sec:model}}
In this section, we describe the model, discuss the different aggregation functions and give details on the training objective.
\subsection {`Row-less' Universal Schema}

While column-less universal schema addresses reasoning over arbitrary textual patterns, it is still limited to reasoning over row entries seen at training time.
\citet{verga2015multilingual} use column-less universal schema for relation extraction.
They address the problem of unseen row entries by using universal schema as a sentence classifier -- directly comparing a textual relation to a KB relation to perform relation extraction.
However, this approach is unsatisfactory for two reasons.
The first is that this creates an inconsistency between training and testing. The model is trained to predict compatibility between rows and columns, but at test time it predicts compatibility between relations directly.
Second, it considers only a single piece of evidence in making its prediction.

\begin{figure}[h]
\centering
\includegraphics[scale=.68]{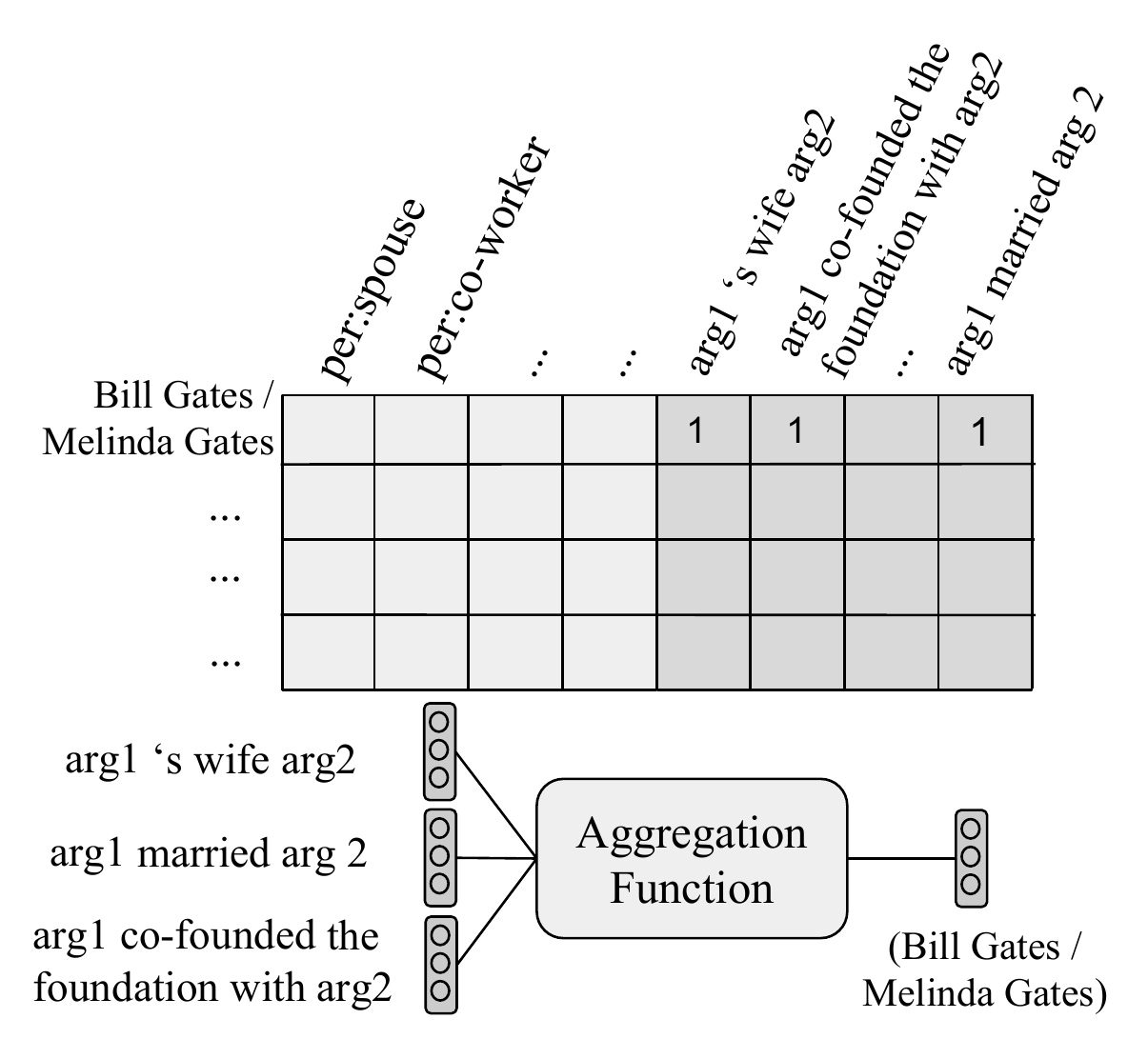}
\caption{Row-less universal schema for relation extraction encodes an entity pair as an aggregation of its observed relation types.
\label{fig:aggregation}}
\end{figure}

We address both of these concerns in our `row-less' universal schema.
Rather than explicitly encoding each row, we encode the row as a learned aggregation over their observed columns (Figure \ref{fig:aggregation}).  A row contains an entity for type prediction and an entity pair for relation extraction while a column contains a relation type for relation extraction and an entity type for type prediction.
A learned row embedding can be seen as a summarization of all  columns observed with that particular row.
Instead of modeling this summarization as a single embedding, we reconstruct a row representation from an aggregate of its column embeddings, essentially learning a mixture model rather than a single centroid. 

\subsection {Aggregation Functions \label{sec:functions}}
In this work we examine four aggregation functions to construct the representations for the row. Let $v(.)$ denote a function that returns the vector representation for rows and columns. To model the probability between row $r$ and column $c$,  we consider the set $\bar{V(r)}$ which contains the set of column entries that are observed with row $r$ at training time, i.e.,

\begin{center}
$\forall \bar{c} \in \bar{V(r)}, (r,\bar{c}) \in T $
\end{center}

The first two aggregation functions create a single representation for each row independent of the query.
\textbf{Mean Pool} creates a single centroid for the row by averaging all of its column vectors, 
\begin{center}
$v(r) = \sum_{\bar{c} \in \bar{V(r)}} v(\bar{c})$
\end{center}
While this formulation intuitively makes sense as an approximation for the explicit row representation, averaging large numbers of embeddings can lead to a noisy representation.

\textbf{Max Pool} also creates a single representation for the row by taking a dimension-wise max over the observed column vectors:
\begin{center}
$v(r)_i = \max_{\bar{c} \in \bar{V(r)}} v(\bar{c})_i, \forall i \in {1,2,\ldots,d}$
\end{center}
where $a_i$ denotes the $i^{th}$ dimension of vector $a$. Both mean pool and max pool are query-independent and form the same representation for the row regardless of the query relation.

We also examine two query-specific aggregation functions.
These models are more expressive than a single vector forced to to act as a centroid to all possible columns observed with that particular row.
For example, the entity pair Bill and Melinda Gates could hold the relation `per:spouse' or `per:co-worker'.
A query-specific aggregation mechanism can produce separate representations for this entity pair dependent on the query.

The \textbf{Max Relation} aggregation function represents the row as its most similar column to the query vector of interest. Given a query relation $c$,
\begin{center}
$c_{max} = argmax_{\bar{c} \in \bar{V(r)}} v(\bar{c}).v(c)$\\
$v(r) = v(c_{max})$
\end{center}
A similar strategy has been successfully applied in previous work~\citep{weston,mssg,neelakantan2015compositional} for different tasks. This model has the advantage of creating a query-specific entity pair representation, but is more susceptible to noisy training data as a single incorrect piece of evidence could be used to form a prediction.

\begin{figure*}[t!]
\hspace{1cm}
\includegraphics[scale=1]{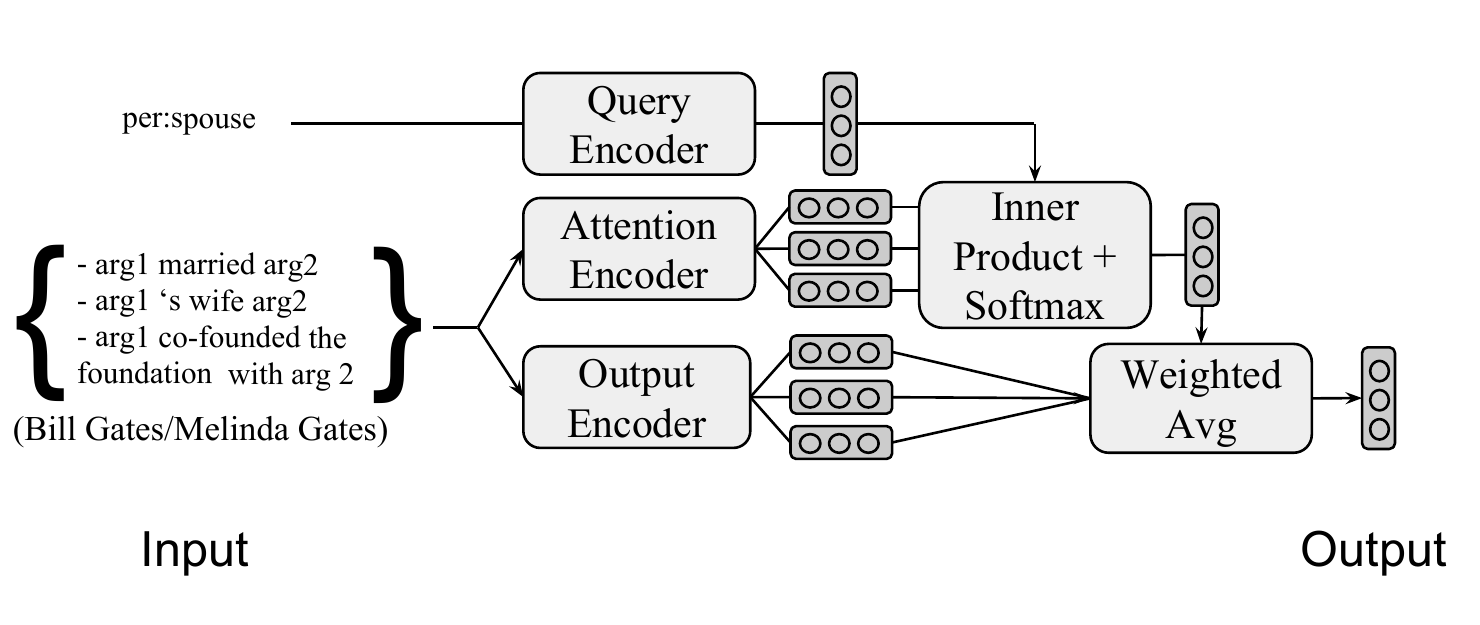}
\caption{\label{fig:attention}
Example attention model in a row-less universal schema relation extractor. In the attention model, we compute the dot product between the representation of the query relation and the representation of an entity pair's observed  relation type followed by a softmax, giving a weighting over the observed relation types.
This output is then used to get a weighted sum over the set of representations of the observed relation types.
The result is a query-specific vector representation of the entity pair.
The Max Relation model takes the most similar observed relation's representation.
}
\end{figure*}

Finally, we look at an \textbf{Attention} aggregation function over columns (Figure \ref{fig:attention}) which is similar to a single-layer memory network \cite{sukhbaatar2015end}. The \emph{soft attention} mechanism has been used to selectively focus on relevant parts in many different models~\citep{Bahdanau2014,GravesWD14,neelakantan2016}.

In this model the query is scored with an input representation of each column embedding followed by a softmax, giving a weighting over each relation type. This output is then used to get a weighted sum over a set of output representations for each column resulting in a query-specific vector representation of the row. Given a query relation $c$,
\begin{center}
$score_{\bar{c}} = v(c).v(\bar{c}), \forall \bar{c} \in \bar{V(r)}$ \\
$p_{\bar{c}} = \frac{exp(score_{\bar{c}})}{\sum_{\hat{c} \in \bar{V(r)}} exp(score_{\hat{c}}) }, \forall \bar{c} \in \bar{V(r)}$ \\
$v(r) = \sum_{\bar{c} \in \bar{V(r)}} p_{\bar{c}} \times v(\bar{c}) $
\end{center}
The model pools relevant information over the entire set of observed columns and selects the most salient aspects to the query.

\begin{table}[h!]
\setlength{\tabcolsep}{4.1pt}
\begin{center}
\begin{tabular}{|l|c|}
\hline
\bf Model & Parameters \\
\hline\hline
Entity Embeddings                & 3.7 e6 \\
Attention                             & 3.1 e5 \\
Mean Pool/Max Pool/Max Relation       & 1.5 e5 \\
\hline
\end{tabular}
\caption{Number of parameters for the different models on the entity type dataset.
\label{table:type_model_parameters}}
\end{center}
\vspace{-.3cm}
\end{table}


\subsection {Training}
The vector representation of the rows and the columns are the parameters of the model. \citet{limin} use Bayesian Personalized Ranking (BPR)~\citep{rendle2009bpr} to train their universal schema models. 
BPR  ranks the probability of observed triples above unobserved triples rather than explicitly modeling unobserved edges as negative.
Each training example is an (entity pair, relation type) or (entity, entity type) pair observed in the training text corpora or KB.

Rather than BPR, \citet{toutanova2015representing} use 200 negative samples to approximate the negative log likelihood\footnote{Many past papers restrict negative samples to be of the same type as the positive example.
We simply sample uniformly from the entire set of row entries}.
In our experiments, we use the sampled approximate negative log likelihood which outperformed BPR in early experiments.

Each example in the training procedure consists of a row-column pair observed in the training set. 
For a positive example $(r,c) \in T$, we construct the set $\bar{V(r)}$ containing all the other column entries apart from $c$ that are observed with row $r$.

To make training faster and more robust, we add `pattern dropout' for entity pairs with many mentions. 
We set $\bar{V(r)}$ to be $m$ randomly sampled mentions for entity pairs with greater than $m$ total mentions.
In our experiments we set $m=10$ and at test time we use all mentions.
We then use $\bar{V(r)}$ to obtain the aggregated row representation as discussed above.

We randomly sample 200 columns unobserved with row $r$ to act as the negative samples.
All models are implemented in Torch\footnote{data and code available at \protect{\url{https://github.com/patverga/torch-relation-extraction/tree/rowless-updates}}} and are trained using Adam \cite{kingma2014adam} with default momentum related hyperparameters.



\section{Related Work}
Relation extraction for KB completion has a long history. \citet{distant_supervision} train per relation linear classifiers using features derived from the sentences in which the entity pair is mentioned. 
Most of the embedding-based methods learn representations for entities \citep{rescal,socherkb,transe} whereas \citet{limin} use entity pair representations. 

`Column-less' versions of Universal Schema have been proposed \citep{toutanova2015representing,verga2015multilingual}. 
These models can generalize to column entries unseen at training by  learning compositional pattern encoders to parameterize the column matrix in place of embeddings. Most of these models do not generalize to unseen entity pairs and none of them generalize to unseen entities. 
Recently, \citet{neelakantan2015compositional} introduced a multi-hop relation extraction model that is `row-less' having no explicit parameters for entity pairs and entities. 

Entity type prediction at the individual sentence level has been studied extensively \citep{pantel,fine-grained,riedel_type}. More recently, embedding-based  methods for knowledge base entity type prediction have been proposed \citep{yao2013universal,neelakantan_type}. These methods have explicit entity representations, hence cannot generalize to unseen entities.

The task of generalizing to unseen row and column entries is referred to as the \emph{cold-start} problem in recommendation systems. Methods proposed to tackle this problem commonly use user and item content and attributes \citep{schein2002methods,cold_start}.

Multi-instance learning can be viewed as the relation classifier analogy of rowless universal schema. 
\citet{riedel2010modeling} used a relaxation of distant supervision training where all sentences for an entity pair (bag) are considered jointly and only the most relevant sentence is treated as the single training example for the bag's label. 
\citet{surdeanu2012multi} extended this idea with multi-instance multi-label learning (MIML) where each entity pair / bag can hold multiple relations / labels. 
Recently \citet{lin2016neural} used a selective attention over sentences in MIML. 

Concurrent to our work, \newcite{row_less} proposes a row-less method for relation extraction considering both a uniform and weighted average aggregation function over columns. 
However, \citet{row_less} did not experiment with max and max-pool aggregation functions or evaluate on entity-type prediction. 
They also did not combine the rowless model with an LSTM column-less parameterization and did not compare to a model with explicit entity-pair representations. 

\section{Experimental Results\label{sec:results}}
In this section, we compare our models that have aggregate row representations with models that have explicit row representations on entity type prediction and relation extraction tasks. Finally, we perform experiments on a column-less universal schema model. Table \ref{table:type_model_parameters} shows that the row-less models require far fewer parameters since they do not explicitly store the row representations.
\subsection{Entity Type Prediction \label{sec:entity-results}}

We first evaluate our models on an entity type prediction task.
We collect all entities along  with their types from a dump of Freebase\footnote{Downloaded March 1, 2015.}.
We then filter all entities with less than five Freebase types leaving a set of $844780$ (entity, type) pairs.
Additionally, we collect $712072$ textual (entity, type) pairs from Clueweb. 
The textual types are the 5000 most common appositives extracted from sentences mentioning entities.
This results in $140513$ unique entities, $1120$ Freebase types, and $5000$ free text types.

All embeddings are $25$ dimensions, randomly initialized.
We tune learning rates from \{.01, .001\}, $\ell_2$ from \{1e-8, 0\}, batch size \{512, 1024, 2048\} and negative samples from \{2, 200\}.

For evaluation, we split the Freebase (entity, type) pairs into $60\%$ train, $20\%$ validation, and $20\%$ test.
We randomly generate $100$ negative (entity, type) pairs for each positive pair in our test set by selecting random entity and type combinations. 
We filter out false negatives that were observed true (entity, type) pairs in our complete data set.
Each model produces a score for each positive and negative (entity, type) pair where the type is the query.
We then rank these predictions, calculate average precision for each of the types in our test set, and then use those scores to calculate mean average precision (MAP).

Table \ref{table:entity_seen_results} shows the results of this experiment. 
We can see that the query dependent aggregation functions (Attention and Max Relation) performs better than the query independent functions (Mean Pool and Max Pool). The performance of models with query dependent aggregation functions which have far fewer parameters match the performance of the model with explicit entity representations.

We additionally evaluate our model's ability to predict types for entities unseen during training.
For this experiment, we randomly select $14000$ entities and take all (entity, type) pairs containing those entities.
We remove these pairs from our training set and use them as positive samples in our test set.
We then select 100 negatives (entity, type) pairs per positive as above.

Table \ref{table:entity_unseen_results} shows the results of the experiment with unseen entities. 
There is very little performance drop for models trained with query dependent aggregation functions.
The performance of the model with explicit entity representations is close to random.

\begin{table}
\centering

\begin{subtable}{.5\linewidth}
{
\begin{tabular}{|l|c|}
\hline
\bf Model & MAP \\
\hline\hline
Entity Embeddings               &  54.81  \\
\hline
Mean Pool                       &  39.47  \\
Max Pool                        &  32.59  \\
\hline
Attention                       & \bf 55.66 \\
Max Relation                    & \bf 55.37 \\
\hline

\end{tabular}
}
\caption{
\label{table:entity_seen_results}}
\end{subtable}%
\vspace{.25cm}
\begin{subtable}{.5\linewidth}
\hspace{.75cm}
{
\begin{tabular}{|l|c|}
\hline
\bf Model & MAP \\
\hline\hline
Entity Embeddings               &  3.14  \\
Mean columns                    &  34.77  \\
Max column                      &  43.20   \\
\hline
Mean Pool                       &  35.53   \\
Max Pool                        &  30.98  \\
\hline
Attention                       & \bf 54.52   \\
Max Relation                    & \bf 54.72   \\
\hline

\end{tabular}
}
\caption{
\label{table:entity_unseen_results}}
\end{subtable}
\caption{Entity type prediction. Entity embeddings refers to the model with explicit row representations. Mean Columns and Max Column are equivalent to Mean Pool and Max Relation respectively (Section \ref {sec:functions}) but use the column embeddings learned during training of the Entity Embeddings model. b: Positive entities are unseen at train time. \label{table:entity_results}}
\end{table}

\begin{table*}[t!]
\small
\begin{tabular}{|l|p{12cm}|}
\hline
\bf Query & Observed Columns \\
\hline\hline
$/$baseball$/$baseball\_player& {\bf $/$sports$/$pro\_athlete}, $/$sports$/$sports\_award\_winner, $/$tv$/$tv\_actor, $/$people$/$measured\_person, $/$award$/$award\_winner, $/$people$/$person \\
$/$architecture$/$engineer &  {\bf engineer},  $/$book$/$author,    $/$projects$/$project\_focus ,  $/$people$/$person   ,      sir \\
$/$baseball$/$baseball\_player & {\bf baseman}, $/$sports$/$pro\_athlete, $/$people$/$measured\_person, $/$people$/$person, dodgers, coach \\
$/$computer$/$computer\_scientist & {\bf $/$education$/$academic}, $/$music$/$group\_member, $/$music$/$artist, $/$people$/$person  \\
$/$business$/$board\_member & {\bf $/$organization$/$organization\_founder}, $/$award$/$award\_winner, $/$computer$/$computer\_scientist, $/$people$/$person, president, scientist \\
$/$education$/$academic & {\bf $/$astronomy$/$astronomer}, $/$book$/$author   \\
\hline

\end{tabular}
\caption{Each row corresponds to a true query entity type (left column) and the observed entity types (right column) for a particular entity. The maximum scoring observed entity type for each query entity type is indicated in bold. The other types are in no particular order. It can be seen that the maximum scoring entity types are interpretable.
\label{table:candies}}
\end{table*}

\subsubsection {Qualitative Results}
A query specific aggregation function is able to pick out relevant columns to form a prediction.
This is particularly important for rows that are not described easily by a single centroid such as an entity with several very different careers or an entity pair with multiple highly varied relations.
For example, in the first row in Table \ref {table:candies}, for the query \emph {$/$baseball$/$baseball\_player} the model needs to correctly focus on aspects like \emph{$/$sports$/$pro\_athlete} and ignore evidence information like \emph{$/$tv$/$tv\_actor}.
A model that creates a single query-independent centroid will be forced to try and merge these disparate pieces of information together.

\subsection{Relation Extraction\label{sec:relation_results}}

We evaluate our models on a relation extraction task using the FB15k-237 dataset from \citet{toutanova2015representing}.
The data is composed of a small set of $237$ Freebase relations and approximately 4 million textual patterns from Clueweb with entities linked to Freebase \cite{gabrilovich2013facc1}.
In past studies, for each (subject, relation, object) test triple, negative examples are generated by replacing the object with all other entities, filtering out triples that are positive in the data set.
The positive triple is then ranked among the negatives.
In our experiments we limit the possible generated negatives to those entity pairs that have textual mentions in our training set.
This way we can evaluate how well the model classifies textual mentions as Freebase relations.
We also filter textual patterns with length greater than $35$.
Our filtered data set contains $2740237$ relation types, $2014429$ entity pairs, and $176476$ tokens.
We report the percentage of positive triples ranked in the top $10$ amongst their negatives as well as the MRR scaled by $100$.

Models are tuned to maximize mean reciprocal rank (MRR) on the validation set with early stopping.
The entity pair model used a batch size $1024$, $\ell_2 = 1e$-$8$, $\epsilon = 1e$-$4$, and learning rate $0.01$.
The aggregation models all used batch size $4096$,  $\ell_2 = 0$, $\epsilon = 1e$-$8$, and learning rate $0.01$.
Each use $200$ negative samples except for max pool which performed better with two negative samples.
The column vectors are initialized with the columns learned by the entity pair model.
Randomly initializing the query encoders and tying the output and attention encoders performed better and all results use this method.
All models are trained with embedding dimension $25$.

Our results are shown in Table \ref{table:seen_results}. 
We can see that the models with query specific aggregation functions give the same results as models with explicit entity pair representations. The Max Relation model performs competitively with the Attention model which is not entirely surprising as it is a simplified version of the Attention model. 
Further, the Attention model reduces to the Max Relation model for entity pairs with only a single observed relation type.
In our data, 64.8\% of entity pairs have only a single observed relation type and 80.9\% have 1 or 2 observed relation types.

We also explore the models' abilities to predict on unseen entity pairs (Table \ref{table:unseen_results}).
We remove all training examples that contain a positive entity pair in either our validation or test set.
We use the same validation and test set as in Table \ref{table:seen_results}.
The entity pair model predicts random relations as it is unable to make predictions on unseen entity pairs. 
The query-independent aggregation functions, mean pool and max pool, perform better than models with explicit entity pair representations. 
Again, query specific aggregation functions get the best results, with the Attention model performing slightly better than the Max Relation model.

The two experiments indicate that we can train relation extraction models without explicit entity pair representations that perform as well as models with explicit representations. We also find that models with query specific aggregation functions accurately predict relations for unseen entity pairs.

\begin{table}
\begin{subtable}{.5\linewidth}
\centering
{
\centering
\begin{tabular}{|l|cc|}
\hline
\bf Model & MRR & Hits@10 \\
\hline\hline
Entity-pair Embeddings          & 31.85 & 51.72 \\
\hline
Mean Pool                       & 25.89 & 45.94 \\
Max Pool                        & 29.61 & 49.93 \\
\hline
Attention                       & 31.92 & 51.67 \\
Max Relation                    & 31.71 & 51.94 \\
\hline
\end{tabular}
}
\caption{
\label{table:seen_results}}
\end{subtable}%
\vspace{.25cm}
\begin{subtable}{.5\linewidth}
\hspace{.75cm}
{
\begin{tabular}{|l|cc|}
\hline
\bf Model & MRR  & Hits@10 \\
\hline\hline
Entity-pair Embeddings          & 5.23          & 11.94        \\
\hline
Mean Pool                       & 18.10         & 35.76       \\
Max Pool                        & 20.80         & 40.25       \\
\hline
Attention                       &\bf 29.75      & 49.69       \\
Max Relation                    & 28.46         & 48.15       \\
\hline
\end{tabular}
}
\caption{
\label{table:unseen_results}}
\end{subtable}
\caption{The percentage of positive triples ranked in the top 10 amongst their negatives as well as the mean reciprocal rank (MRR) scaled by 100 on a subset of the FB15K-237 dataset. All positive entity pairs in the evaluation set are unseen at train time. Entity-pair embeddings refers to the model with explicit row representations. b: Predicting entity pairs that are not seen at train time.\label{table:relex}}
\end{table}

\subsection {`Column-less' universal schema}

The original universal schema approach has two main drawbacks: similar textual patterns do not share statistics, and the model is unable to make predictions about entities and textual patterns not explicitly seen at train time.

Recently, `column-less' versions of universal schema  to address some of these issues \citep{toutanova2015representing,verga2015multilingual}.
These models learn compositional pattern encoders to parameterize the column matrix in place of direct embeddings.
Compositional universal schema facilitates more compact sharing of statistics by composing similar patterns from the same sequence of word embeddings -- the text patterns `lives in the city' and `lives in the city of' no longer exist as distinct atomic units.
More importantly, compositional universal schema can thus generalize to all possible textual patterns, facilitating reasoning over arbitrary text at test time.

%

The column-less universal schema model generalizes to all possible input textual relations and the row-less model generalizes to all entities and entity pairs, whether seen at train time or not. We can combine these two approaches together to make an universal schema model that generalizes to unseen rows and columns.

The parse path between the two entities in the sentence is encoded with an LSTM model. We use a single layer model with $100$ dimensional token embeddings initialized randomly.  To prevent exploding gradients, we clip them to norm $10$ while all the other hyperparameters are tuned the same way as before. We follow the same evaluation protocol from \ref{sec:relation_results}.

The results of this experiment with observed rows are shown in Table \ref {table:columnless_seen_results}. While both the MRR and Hits@10 metrics increase for models with explicit row representations, the row-less models show an improvement only on the Hits@10 metric. The MRR of the query dependent row-less models is still competitive with the model with explicit row representation even though they have far fewer parameters to fit the data.

\begin{table}

\begin{subtable}{.5\linewidth}
\centering
{
\begin{tabular}{|l|cc|}

\hline
\bf Model & MRR & Hits@10 \\
\hline\hline
Entity-pair Embeddings          & 31.85 & 51.72 \\
Entity-pair Embeddings-LSTM     &\bf 33.37 & 54.39 \\
\hline
Attention                       & 31.92 & 51.67 \\
Attention-LSTM                  & 30.00 & 53.35 \\
\hline
Max Relation                    & 31.71 & 51.94 \\
Max Relation-LSTM               & 30.77 &\bf 54.80 \\
\hline
\end{tabular}
}
\caption{
\label{table:columnless_seen_results}}
\end{subtable}%
\vspace{.1cm}
\begin{subtable}{.5\linewidth}
\hspace{.75cm}
{
\begin{tabular}{|l|cc|}
\hline
\bf Model & MRR  & Hits@10 \\
\hline\hline
Entity-pair Embeddings          & 5.23          & 11.94        \\
\hline
Attention                       &\bf 29.75      & 49.69       \\
Attention-LSTM                  & 27.95         & 51.05     \\
\hline
Max Relation                    & 28.46         & 48.15       \\
Max Relation-LSTM               & 29.61         &\bf 54.19  \\
\hline
\end{tabular}
}
\caption{
\label{table:columnless_unseen_results}}
\end{subtable}
\caption{The percentage of positive triples ranked in the top 10 amongst their negatives as well as the mean reciprocal rank (MRR) scaled by 100 on a subset of the FB15K-237 dataset. Negative examples are restricted to entity pairs that occurred in the KB or text portion of the training set. Models with the suffix ``-LSTM'' are column-less. Entity-pair embeddings refers to the model with explicit row representations. b: Predicting entity pairs that are not seen at train time.\label{table:columnless}}
\end{table}

\section{Conclusion}
In this paper we explore a row-less extension of universal schema that forgoes explicit row representations for an aggregation function over its observed columns.
This extension allows prediction between all rows in new textual mentions -- whether seen at train time or not -- and also provides a natural connection to the provenance supporting the prediction. Our models also have a smaller memory footprint.

In this work we show that an aggregation function based on query-specific attention over relation types outperforms query independent aggregations.
We show that aggregation models are able to predict on par with models with explicit row representations on seen row entries with far fewer parameters. More importantly, aggregation models predict on unseen row entries without much loss in accuracy. 
Finally, we show that in relation extraction, we can combine row-less and column-less models to train models that generalize to both unseen rows and columns.

\subsubsection*{Acknowledgments}
We thank Emma Strubell, David Belanger, and Luke Vilnis for helpful discussions and edits.
This work was supported in part by the Center for Intelligent Information Retrieval and the Center for Data Science, and in part by DARPA under agreement number FA8750-13-2-0020. The U.S. Government is authorized to reproduce and distribute reprints for Governmental purposes notwithstanding any copyright notation thereon, in part by Defense Advanced Research Agency (DARPA) contract number HR0011-15-2-0036, and in part by the National Science Foundation (NSF) grant number IIS-1514053. Any opinions, findings and conclusions or recommendations expressed in this material are those of the authors and do not necessarily reflect those of the sponsor. Arvind Neelakantan is
supported by a Google PhD fellowship in machine learning.

\bibliography{sources}
\bibliographystyle{eacl2017}

\newpage

\end{document}